\documentclass{article}
\usepackage{spconf}
\usepackage{mathptmx}
\usepackage{amsmath}
\usepackage{amssymb}
\usepackage{graphicx}
\usepackage{booktabs}
\usepackage[table]{xcolor}
\definecolor{softgreen}{RGB}{233,245,236}
\definecolor{darkgreen}{RGB}{27,94,32}
\usepackage[T1]{fontenc}
\usepackage[protrusion=true,expansion=true,tracking=false]{microtype}
\usepackage{balance}
\usepackage{comment} 
\graphicspath{{figs/}{figs/w32/}{tables/}{tables/w30/}{tables/w32/}{tables/w34/}}


\makeatletter
\def\thmlist{\list{}{\topsep 1pt \partopsep 0pt \itemsep 0pt \parsep 0pt
  \leftmargin\z@ \rightmargin\z@ \listparindent\z@ \itemindent\z@}}
\def\@begintheorem#1#2{\thmlist\item\relax{\bf #1\ #2.}\hskip 0.5em\it\ignorespaces}
\def\@opargbegintheorem#1#2#3{\thmlist\item\relax{\bf #1\ #2\ (#3).}%
  \it\ignorespaces}
\def\@endtheorem{\endlist}
\makeatother
\newtheorem{proposition}{Proposition}

\makeatletter
\renewcommand\section{\@startsection{section}{1}{\z@}%
  {-1.2ex plus -0.3ex minus -.2ex}{0.4ex plus .2ex}{\normalsize\bf}}
\renewcommand\subsection{\@startsection{subsection}{2}{\z@}%
  {-1.0ex plus -0.3ex minus -.2ex}{0.25ex plus .2ex}{\normalsize\bf}}
\renewcommand\paragraph[1]{\par\addvspace{0.85ex}\noindent
  {\normalsize\bf #1}\hskip 0.5em\ignorespaces}
\def\captionfont{\fontsize{9}{10}\selectfont}

\long\def\@makecaption#1#2{\vskip 3pt
 \setbox\@tempboxa\hbox{\captionfont #1. #2}
 \ifdim \wd\@tempboxa >\hsize
   {\captionfont #1. #2\par}
 \else
   \hbox to\hsize{\hfil\box\@tempboxa\hfil}\fi}
\makeatother
\makeatletter
\def\thebibliography#1{\section{References}\list
 {[\arabic{enumi}]}{\settowidth\labelwidth{[#1]}\leftmargin\labelwidth
 \advance\leftmargin\labelsep \itemsep 0pt \parsep 0pt \topsep 0pt
 \usecounter{enumi}}
 \def\newblock{\hskip .11em plus .33em minus .07em}
 \sloppy\clubpenalty4000\widowpenalty4000
 \sfcode`\.=1000\relax}
\makeatother
\makeatletter
\def\tightlist{\list{$\bullet$}{\topsep 1pt \partopsep 0pt \itemsep 1pt
  \parsep 0pt \leftmargin 1.1em \labelwidth 0.7em \labelsep 0.4em
  \rightmargin\z@}}

\def\gloss#1{\par\vskip 1pt\noindent{\fontsize{9}{10}\selectfont #1}\par\vskip 1pt}
\makeatother

\title{Robust Explanatory Analysis for Structured Optimization Networks (REASON) in Learning}

\name{Vahid Jebraeeli$^{1}$, Hamid Krim$^{1}$, Keisuke Fujii$^{2}$}
\address{$^{1}$ECE Department, North Carolina State University, Raleigh, NC, USA \\
         $^{2}$Graduate School of Informatics, Nagoya University, Nagoya, Japan \\
         \texttt{vjebrae@ncsu.edu} \qquad \texttt{ahk@ncsu.edu} \qquad \texttt{fujii@i.nagoya-u.ac.jp}}

\begin{document}
\ninept
\maketitle
\renewcommand{\thefootnote}{\arabic{footnote}}
\setcounter{footnote}{0}

\begin{abstract}
We seek a hierarchical resolution of residual error under a composite learning objective as a principled mechanism for identifying and exploiting additional information, particularly under distribution shift and perturbation. Inspired by the System~1/System~2 paradigm, we formulate this principle as a two-loop mechanism serving a common inference goal. A fast, System~1-like loop resolves the residual addressable by the available evidence; the remaining epistemic residual triggers a medium speed, System~2-like loop that actively acquires additional evidence and measures its informational value through residual reduction. Over nested solution spaces, the medium-loop optimum is provably no worse than the fast-loop optimum. Experiments show that the resulting two-loop model consistently improves upon the fast loop at approximately half the computation of the medium loop, including under unseen corruptions, and, when provided a cleaner observation, outperforms staged classifiers and learning-to-defer methods at equal computational cost.
\end{abstract}


\section{Introduction}
\label{sec:intro}
While machine learning has achieved remarkable success, much of this progress has come at the expense of increasing model and sample complexity while still falling short on generalization. Transformers have partially advanced beyond this paradigm by explicitly modeling cross-contextual coherence through attention\cite{vaswani2017attention}, thereby enabling a modest but important form of relational reasoning. Yet this reasoning remains largely associative and compositional, rather than reflecting a deeper structural organization of memory, inference, and adaptive computation.
The notion of reasoning has long been regarded as an inseparable component of intelligent learning, yet its principled and coherent integration into learning systems has remained elusive. A major conceptual breakthrough emerged from the work of Kahneman and Tversky\cite{kahneman2011thinking}, and the subsequent fast–slow, or System~1/System~2, view of cognition, which provided a simple yet powerful framework for rationalizing how learning and decision-making may operate across distinct computational regimes: a fast, intuitive, and largely automatic process, complemented by a slower, deliberative, and reasoning-intensive process. Complementarily, the consciousness-prior perspective \cite{bengio2017consciousness} suggests that such higher-level processing may operate on a selectively filtered subset of the available representation, with attention providing a natural mechanism for funneling task-relevant information toward more deliberate reasoning. 
In a different but related way, we propose learning as a hierarchy of loops that resolve residual uncertainty over progressively richer solution spaces. The Fast Loop (FL) resolves errors addressable by the current amortized inference map, while persistent residuals activate a Medium Speed Loop (MSL) that explores alternative reasoning trajectories without altering the underlying representation. Failure of this expanded search to sufficiently reduce the residual is then interpreted as evidence of a representational limitation, ideally triggering a creative loop that operates over an architectural state space to adapt the hypothesis class itself, a third level we defer to future work. 
Our proposed approach shares with selective classification, failure detection, learning-to-defer, and adaptive-inference methods, the recognition at a high level that computational effort should be non-uniform across samples. Classical reject-option and selective-prediction methods determine whether an inference should be accepted or withheld \cite{chow1970optimum,geifman2017selective}, while failure-detection and calibration methods seek reliable indicators of when a prediction is likely to be erroneous or out of distribution \cite{hendrycks2017baseline,guo2017calibration,jaeger2023call}. Learning-to-defer methods extend this principle by routing uncertain decisions to a downstream expert or model \cite{mozannar2020consistent,narasimhan2022posthoc,jitkrittum2023confidence}, whereas staged and early-exit architectures adapt computational cost by allowing easy examples to terminate early and difficult examples to invoke additional processing \cite{viola2001rapid,teerapittayanon2016branchynet,bolukbasi2017adaptive}. Our formulation views failure beyond the criterion for rejection, deferral, or additional computation, but as a structured residual which  determines the type of computation subsequently activated. 
Thus, rather than implementing a confidence-based sequence of predictors of increasing complexity, the proposed hierarchy progressively expands the admissible modes of inference, reasoning, and representation according to the nature of the unresolved residual.

To sum up and in contrast to fore-noted existing work hinging on pre-selected gating mechanism as well as to the Bayesian interpretation of System~1/System~2 in \cite{Friston2010}, the central objective of this paper is the realization of a data-driven learning system inspired by the latter paradigm and built around a two-loop learning architecture (Fast/Medium) resulting from a hierarchical optimization formulation of a learning functional objective. Our implementation also refers inputs by a calibrated confidence threshold; what differs from prior gating is the action it triggers, the acquisition of new evidence. 

Two key questions are addressed in this contribution. First, can decision-making be improved by selectively acquiring additional evidence to reduce the residual error, rather than through a uniformly optimized decision rule of comparable model complexity governed by a preselected threshold gate? This directly examines the learning value of the System~1/System~2 paradigm, wherein additional computational and informational resources are invoked only when the fast-loop inference leaves a significant unresolved residual, thereby allocating learning and reasoning resources where they are most needed.

Second, what is the value of the MSL operating in \textbf{REVEAL} mode, where additional evidence is actively sought to resolve uncertainty remaining after fast-loop inference? The utility of the acquired evidence is measured by its informational novelty relative to the current representation and, critically, by its reduction of the epistemic residual $R_{\mathrm{epi}}$. It is thus assessed not by the mere acquisition of additional evidence, but by its ability to acquire informative evidence that measurably resolves epistemic uncertainty.
 Extensive experiments are conducted across four fine-grained recognition benchmarks: Stanford Dogs, CUB-200-2011, FGVC-Aircraft, and Oxford-IIIT Pets.\par
 Here, we shift focus from efficient learning with fixed data availability\cite{jebraeeli2024koopcon,jebraeeli2025expansive} to inference, where computational and informational resources are rather adaptively allocated on a per-input basis. The remainder of the paper formulates the problem and develops the proposed solution, presents the experimental results in Section 4, and concludes in Section~5.
\section{Problem Statement and Solution}
\label{sec:system}
\subsection{Residual decomposition over nested regimes}
\label{sec:residual}
We formulate learning as the hierarchical resolution of residual uncertainty over progressively richer solution spaces. A FL, instantiated as a conventional CNN, is trained to minimize the inference residual $R_{\mathrm{inf}}$ associated with the current inference map. Persistent residual uncertainty $R_{\mathrm{epi}}$ activates a MSL, which, upon request from the FL, explores alternative decisions using additional evidence. As noted earlier, the natural triggering of a creative loop (CL) upon failure of these two procedures is deferred to future work. Accordingly, the associated architecture state space may be defined as
\[\Omega=\mathcal{C}\times\mathcal{L}\times\mathcal{P},\]
where $\mathcal{C}$, $\mathcal{L}$, and $\mathcal{P}$ respectively parameterize the search over channel capacity, network depth, and interaction complexity/order.
A complete state is defined as  $X = (z,\theta,\psi)$, with $z$ the latent
inference state of an observation, $\theta$ the fast predictor which consults  $\psi$ the
predictive model as the medium regime. The residual of a decision
decomposes approximately into a part the current inference map can already
resolve and a part left by insufficient evidence,
\begin{equation}
R \;\approx\; R_{\mathrm{inf}} + R_{\mathrm{epi}},
\label{eq:residual}
\end{equation}
and the two regimes target the two parts: the FL $\mathcal{T}_f$
reduces $R_{\mathrm{inf}}$ by amortized inference, and the MSL
$\mathcal{T}_m$ acts only on what it leaves. A complete sweep is therefore a
composition,
\begin{equation}
X^{(k+1)} = \mathcal{T}_m \circ \mathcal{T}_f\bigl(X^{(k)}\bigr),
\qquad
\mathcal{T}_m \text{ active iff } R_{\mathrm{epi}} > \varepsilon_m,
\label{eq:sweep}
\end{equation}
in which the medium speed loop is bypassed and the fast solution retained wherever
the epistemic residual is already small. The organization is a nonlinear
block-coordinate relaxation: distinct mechanisms address distinct components
of one residual, and the slower one is invoked only for what the faster one
cannot resolve. 

Both regimes descend a single functional,
\begin{equation}
\mathcal{J}(X) = \mathcal{L}_{\mathrm{task}} + \lambda_p
\mathcal{L}_{\mathrm{pred}},
\label{eq:objective}
\end{equation}
encoding task performance and predictive consistency, over hierarchically
nested feasible sets $S_F \subset S_M$: in $S_F$ the latent state is
what the fast map returns, while $S_M$ also admits a state revised on
further evidence. Nesting alone then gives the following.
\begin{proposition}  
Consider a two-loop learning system consisting of a fast loop $F$ and a medium speed loop $M$, operating over hierarchically nested admissible solution spaces
$S_F\subseteq S_M$.
Let X denote an admissible learning state, including the latent representation and the parameters active at the corresponding loop, and let both loops be governed by the common objective
\[\mathcal J(X)
=
\mathcal L_{\mathrm{task}}(X)
+
\lambda_p\mathcal L_{\mathrm{pred}}(X),
\qquad \lambda_p\geq0.\]
Assume that a minimizer of $\mathcal J$ exists over each admissible space. Define
$X_F^\star
=
\arg\min_{X\in S_F}\mathcal J(X),
\qquad
X_M^\star
=
\arg\min_{X\in S_M}\mathcal J(X).$ Then\\

${
\mathcal J(X_M^\star)
=
\min_{X\in S_M}\mathcal J(X)
\leq
\min_{X\in S_F}\mathcal J(X)
=
\mathcal J(X_F^\star).
}$
\end{proposition}

Sketch of Proof: Since $S_F\subseteq S_M$, the fast-loop optimum $X_F^\star$ is feasible in $S_M$. Hence, minimizing the same objective over the enlarged space yields
$\mathcal J(X_M^\star)\leq \mathcal J(X_F^\star).$
\hfill\rule{1mm}{1mm}\\

The inequality establishes that the optimal medium-regime performance is no worse than that of the fast regime and can improve when additional evidence is effectively exploited. In the experiments, however, $\theta$ and $\psi$ are trained separately, each on its own objective rather than by minimizing $\mathcal J$ jointly, and then held fixed; only the latent state $z$ varies per input, and medium speed loop activation is governed by a threshold rather than direct minimization. The proposition therefore motivates the design but does not guarantee a gain for the implemented gate; the relevant empirical questions are when the inequality is strict, the magnitude of the improvement, and its computational cost.

\subsection{Fast Loop and Epistemic Action}
\label{sec:fast}
With fixed parameters, the FL maps each observation to a decision,
\begin{equation}
s=\Phi_\theta(o), \qquad
\hat y_1=\arg\max_y\pi_\theta(y\mid s), \qquad
c(s)=\max_y\pi_\theta(y\mid s),
\end{equation}
where \gloss{$o$: observation; $s$: feature representation; $\pi_\theta$: FL decision distribution; $c(s)$: confidence.}

The epistemic residual $R_{\mathrm{epi}}$ is assessed through confidence, with low confidence triggering the MSL:
\begin{equation}
c(s)<c_0, \qquad
c_0=Q_\gamma(c),
\label{eq:admit}
\end{equation}
where \gloss{$Q_\gamma$: the $\gamma$-quantile of pooled calibration confidence, with $\gamma$ the low-confidence fraction.}

Thus, $c_0$ operationalizes $\varepsilon_m$, with low confidence indicating a large unresolved residual. Its quantile definition provides a consistent low-confidence fraction across benchmarks. The sole parameter $\gamma$ is set on the calibration set, never on the test set.

\subsection{The second observation}
\label{sec:second}
\begin{figure}[t]\centering
\includegraphics[width=\columnwidth]{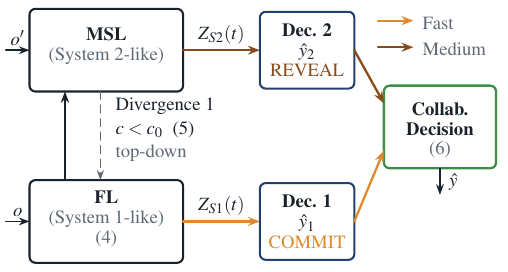}
\caption{The FL processes each input and either adopts its decision (COMMIT) or, when confidence is insufficient, invokes the MSL to acquire and process additional evidence (REVEAL).}
\label{fig:arch}
\end{figure}

A COMMIT action preserves the FL decision, whereas REVEAL acquires supplemental evidence $o'$ for the medium speed loop, as illustrated in Fig.~\ref{fig:arch}:
\begin{equation}
\hat y=
\begin{cases}
\hat y_2=\arg\max_y \pi_\psi(y\mid o'), & \text{REVEAL},\\
\hat y_1, & \text{COMMIT},
\end{cases}
\label{eq:reveal}
\end{equation}
where \gloss{$\pi_\psi$: fixed-parameter MSL decision distribution; $o'$: supplemental evidence, comprising either a degraded sample, an independent draw under the same corruption, or a clean view.}

The FL incurs cost $c_1$, while each REVEAL incurs an additional MSL cost $c_2$. Thus, at low-confidence fraction $\gamma$, the average cost per input is
\begin{equation}
\bar c(\gamma)=c_1+\gamma c_2,
\label{eq:cost}
\end{equation}
where \gloss{$\gamma$: low-confidence fraction in Eq.~(\ref{eq:admit}), equivalently the referral rate. Cost is measured in GMACs ($10^9$ multiply-accumulate operations); evidence acquisition is assigned zero cost.}
The three cases provide objective benchmarks for evidence quality: the clean view establishes an upper bound, reprocessing the degraded input provides a baseline, and an independent draw provides fresh evidence of the same quality.

At a set $\gamma$, all three options 
incur the same cost, so that any difference between them is attributable to the observation alone.

\section{Experimental Design}
\label{sec:design}
\paragraph{Benchmarks.} Four fine-grained benchmarks are used: Stanford Dogs \cite{khosla2011dogs},
CUB-200-2011 \cite{wah2011cub}, FGVC-Aircraft \cite{maji2013aircraft}, and Oxford-IIIT Pets
\cite{parkhi2012pets} (Dogs, CUB, Aircraft, and Pets). Dogs is split per class into $14{,}355$
training, $3{,}025$ calibration, and $3{,}200$ test images; its accuracies are therefore not
comparable to published results. Its images are also drawn from ImageNet~\cite{deng2009imagenet}, which is part of the
pretraining data of both backbones, so its absolute accuracies may be optimistic, although all
compared models share this pretraining. The other three benchmarks retain their official test sets, with
calibration sets of $1{,}200$, $1{,}300$, and $736$ images held out from their training data, on
which every operating point is selected and which no FL has seen.

\paragraph{Models.} The FL is MobileNetV3-Large \cite{howard2019mobilenetv3} at $224 \times 224$ px (pixels), and the
MSL is DINOv2-B \cite{oquab2024dinov2} at $518$ px, adapted with LoRA \cite{hu2022lora} and
classifying by distance to class exemplars; both are fine-tuned once and then held fixed. The initial
CUB and Aircraft FLs fell short of the published accuracy of their backbone and were therefore
retrained with a procedure selected on the calibration set, each test set being evaluated
only once: clean accuracy rose from $68.31$ to $81.72$ on CUB and from $58.54$ to $87.70$ on
Aircraft, against published values of $73.91$ and $85.8$
\cite{jeevan2024backbone,kornblith2019better}. The Dogs and Pets FLs are retained as
initially trained: the Pets FL already exceeds its published value, and the Dogs FL is $5.1$ points
below a linear classifier on its frozen backbone ($1.5$ for Pets).
The single model of Section~\ref{sec:results} is the same MSL
retrained at $392$ px under the identical protocol.

\paragraph{Conditions.} In addition to the clean images, seven perturbations of our own
construction are applied, following \cite{hendrycks2019benchmarking}: occlusion at three sizes,
cutout, feature-region masking, additive noise, and blur. Unless stated otherwise, each reported accuracy is the mean over these eight conditions, with
the MSL reprocessing the degraded input.

\paragraph{Equal cost.} The three cases refer the same inputs in each condition and hence incur the
same cost. The FL requires $c_1 = 0.22$ GMAC per input ($0.29$ on CUB) and the MSL $c_2 = 154.94$
GMAC.

\paragraph{The operating point.} The threshold $c_0$ is chosen so that the MSL processes the half of the inputs for which the FL is least confident, i.e., $\gamma=1/2$. This operating point is selected on the calibration set to maximize the gain over the FL while referring no more than half of the inputs to the MSL.

\paragraph{Training and testing.} The FL and MSL are trained separately on each benchmark’s training set, including images degraded by perturbations of the same types as the test conditions, and are then frozen. Using held-out calibration images never seen during training, $c_0$ is set to the median FL confidence across all eight conditions. At test time, the FL first classifies every image: if its confidence is at least $c_0$, the decision is retained (COMMIT); otherwise, the MSL processes a second observation of the same object (the same degraded image, an independent draw of the same perturbation, or a clean view), and its decision is used instead (REVEAL). Generalization beyond the eight conditions is evaluated by holding each condition out when setting $c_0$ and by testing on fifteen corruption types unseen during training (Section~\ref{sec:heldout}). Performance differences are reported with $95$ percent confidence intervals over test images.

\section{Results}
\label{sec:results}
Sections~\ref{sec:vs1} and~\ref{sec:reacq} address the two questions of Section~\ref{sec:intro} in turn, and
Section~\ref{sec:heldout} tests generalization.
\subsection{Accuracy and Cost}
\label{sec:vs1}
\begin{figure}[t]\centering
\includegraphics[width=\columnwidth]{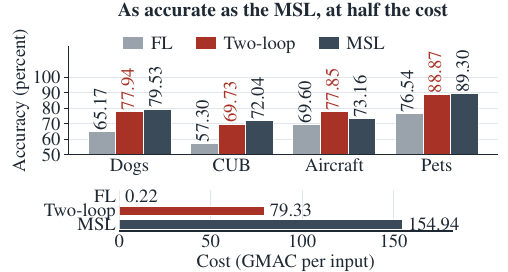}
\caption{The FL alone, the two-loop model (FL $+$ MSL), and the MSL alone,
means over eight conditions (top), cost per input (bottom).}
\label{tab:main}
\end{figure}
\begin{figure}[t]\centering
\includegraphics[width=\columnwidth]{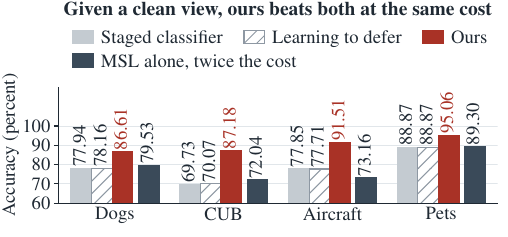}
\caption{Staged classifiers~\cite{viola2001rapid} and learning to defer (best of
three post-hoc rules~\cite{narasimhan2022posthoc,jitkrittum2023confidence}),
run with our FL and MSL at the same cost of about $78$ GMAC per input,
reprocess the input the FL saw, as in Fig.~\ref{tab:main}; ours receives a clean view instead, so the
gap measures the value of better evidence, acquisition cost excluded.
Means over eight conditions.}
\label{fig:compare}
\end{figure}
Fig.~\ref{tab:main} sets the FL, the two-loop model (FL $+$ MSL)
and the MSL side by side, with their respective cost; every model processes the degraded input.

\paragraph{Analysis.} The two-loop model improves on the FL on every benchmark, by $8$–$13$ points
when the MSL reprocesses the degraded input and by $19$–$30$ points when it processes a clean view.
With the degraded input, it comes close to the MSL alone on three benchmarks and exceeds it on
Aircraft, at roughly half its cost (Fig.~\ref{tab:main}). The confidence ranking accounts for
less than half of this gain: against a random referral of the same size, it adds $4$–$5$ points
with the degraded input and $5$–$7$ with a clean view.

\paragraph{A single model at the same cost.} Retraining the MSL at $392$ px, with all else
unchanged, yields a single model requiring $80.31$ GMAC per input, slightly more than the two-loop model
on three benchmarks. On the degraded input, it outperforms the two-loop model on Dogs, CUB, and Pets by
$1.3$–$2.6$ points and underperforms it on Aircraft by $3.6$ points, all four differences being
significant.

\paragraph{Choosing between the two designs.} The two-loop model assigns the more confident half of the inputs to the FL alone, reducing computation without sacrificing accuracy only when the FL is at least as accurate as the MSL on these retained inputs. On the calibration set, the FL exceeds the MSL on this subset by $9.4$ points for Aircraft but underperforms it on the other three benchmarks; the test sets preserve these signs within $0.5$ points. Because the two-loop model and MSL make identical decisions on the referred fraction $\gamma$, their accuracy difference is exactly
$[\operatorname{acc}(\mathrm{FL}\mid\mathrm{kept})-\operatorname{acc}(\mathrm{MSL}\mid\mathrm{kept})](1-\gamma).$
Thus, since the single model exceeds the MSL by $0.3$–$1.0$ points on every benchmark, a negative gap favors the single model, whereas the large positive gap on Aircraft allows the two-loop model to prevail.

\paragraph{Comparison with staged classifiers and learning-to-defer.}
The distinction is not the gate, but the action it triggers. At the same referral rate, entropy, energy, margin, and temperature-scaled confidence are within $1.4$ points of max-softmax, while the best post-hoc learned deferral rule~\cite{narasimhan2022posthoc,jitkrittum2023confidence} is within $0.4$ points~\cite{jaeger2023call}. Staged classifiers~\cite{viola2001rapid}, early exit~\cite{teerapittayanon2016branchynet}, and learning-to-defer reprocess the available input; REVEAL instead permits acquisition of new evidence. At matched cost, a clean view improves over reprocessing by $6$–$17$ points and over the MSL alone by $6$–$18$ points at half its cost (Fig.~\ref{fig:compare}), whereas an independent draw under the same corruption yields no gain (Section~\ref{sec:reacq}).
\subsection{Gain of Supplemental Evidence} 
\label{sec:reacq}
The same two-loop model is run three times at the same operating point and cost, changing only the
evidence supplied to the MSL: the degraded input processed by the FL, an independent draw of the
same corruption at the same severity, or a clean view. Since the referred inputs are
identical across the three runs, any difference is attributable to the observation alone.

\paragraph{Analysis.} 
A clean view gains $6$–$17$ points over reprocessing across the eight conditions and $2$–$4$ on the three described next. For cutout, noise, and blur, an independent draw is simply a fresh sample from the same corruption distribution: for the MSL applied to every input, it changes accuracy by $-0.7$ to $0.0$ points relative to reprocessing, with two intervals below zero and none above; for the two-loop model, the change is within $0.1$ points of zero, with every interval straddling zero. Thus, acquisition benefits primarily from improved evidence quality, not novelty.

Feature masking is the exception. Since each draw masks one of five regions, redrawing leaves aggregate FL performance essentially unchanged (within $0.6$ points), but thresholding selects cases in which the discriminative region was masked. A fresh draw then improves the FL by $3$–$11$ points and the two-loop model by $0.2$–$2.5$ points (significant only on CUB). A clean view still gains $2$–$5$ points, at least twice the redraw gain. The three graded occlusions are excluded because their initial occluder is centered whereas the redraw is uniform, creating a more favorable distribution under which the FL alone gains $2$–$6$ points.

Finally, REVEAL replaces rather than fuses observations. An oracle selecting, per image, the better of reprocessing and an independent draw would raise the MSL's accuracy by $1$–$4$ points over reprocessing; the present design cannot attain this bound because the MSL does not retain the first observation.
\subsection{Generalization}
\label{sec:heldout}
\paragraph{Held-out conditions.} Generalization is assessed by leaving one condition out at a time
\cite{gulrajani2021search,chollet2019measure}: the threshold $c_0$ is set to the median calibration
confidence over the remaining seven conditions, and the two-loop model is tested on the held-out one.
On every held-out condition, the two-loop model remains above the FL,
on average by $8$–$13$ points per benchmark when reprocessing the degraded input and by $19$–$30$
points with a clean view.

\begin{figure}[t]\centering
\includegraphics[width=\columnwidth]{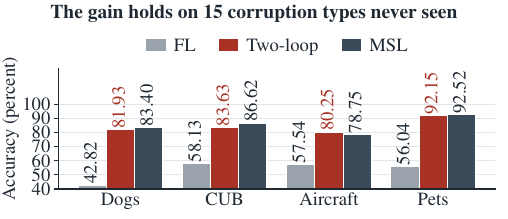}
\caption{The FL alone, our two-loop model, and the MSL alone on the fifteen
ImageNet-C corruption types~\cite{hendrycks2019benchmarking} at severity three,
unseen in training and calibration.}
\label{fig:heldout}
\end{figure}
\paragraph{Unseen corruptions.} Each benchmark's threshold $c_0$ is then applied, unchanged, to the
fifteen ImageNet-C corruption types \cite{hendrycks2019benchmarking} at severity three, generated with
the implementation of~\cite{michaelis2019benchmarking}. The two-loop
model outperforms the FL on every benchmark and corruption type (Fig.~\ref{fig:heldout}); it is within three points of the MSL alone on every
benchmark and above it on Aircraft, at $58$–$79$ percent of its cost. The referral rate is not
prescribed but follows from the threshold: on Dogs it ranges from $30$ percent of the inputs under
brightness to all of them under Gaussian noise, where the FL accuracy falls below one percent and
the two-loop model then requires more computation than the MSL alone. Here too, an independent
draw yields no gain.

\section{Discussion and Conclusion}
\label{sec:conc}
\paragraph{Two-Loop Model: Strengths and Limitations.} With a fixed FL–MSL pair, confidence-thresholded referral, and a second observation acquired only when the FL is uncertain, the two-loop model gains $8$–$13$ points over the FL at roughly half the MSL cost. The gain persists across all held-out conditions and an unseen corruption benchmark, exceeds random referral at matched cost, and satisfies the inequality of Section~\ref{sec:residual} on every benchmark. However, retraining the MSL at the computational cost of the two-loop model performs better on three of four benchmarks. The exception is precisely where the FL classifies its retained inputs more accurately than the MSL, a condition detectable from calibration before training the alternative. Moreover, the benefit of the second observation arises primarily from improved quality rather than novelty. Since Eq.~(\ref{eq:cost}) assigns zero cost to evidence acquisition, selective acquisition is advantageous only when acquiring evidence itself carries a cost, as when a second viewpoint, a higher-resolution or longer-exposure capture, or a retake must be requested; otherwise, a single model of the same computational cost is preferable.

\paragraph{Running time and alternatives.} In running time rather than arithmetic
operations, the reduction is smaller: about $70$ percent of the MSL's time on Dogs, rather than
half. Given the same clean view, the FL alone still trails the two-loop model by $3.8$–$5.5$ points.

\paragraph{Future work.} Two directions remain open: fusing both observations, and an
epistemic-residual test that improves upon confidence.

\clearpage
\section{Compliance with Ethical Standards}
This study involved no human or animal subjects. It used four publicly available benchmarks of
photographs of dogs, birds, aircraft, and pets~\cite{khosla2011dogs,wah2011cub,maji2013aircraft,parkhi2012pets},
the Stanford Dogs images being drawn from ImageNet~\cite{deng2009imagenet}, each used for
non-commercial research under its license or terms of use, together with publicly released
pretrained models~\cite{howard2019mobilenetv3,oquab2024dinov2}. No data were collected and no live
animals were involved. Some photographs incidentally show people; no individual was identified or
analyzed, and no dataset images are reproduced in this paper. Ethical approval was therefore not
required.

\section{Acknowledgments}
This work was supported by the College of Engineering, North Carolina State University, and the
Graduate School of Informatics, Nagoya University. The authors have no relevant financial or
non-financial interests to disclose.

\balance
\bibliographystyle{IEEEbib}
\bibliography{refs,refs_comparison,refs_repositioning}

\begin{thebibliography}{10}

\bibitem{vaswani2017attention}
Ashish Vaswani et~al.,
\newblock ``Attention is all you need,''
\newblock in {\em Advances in Neural Information Processing Systems (NeurIPS)}, 2017, pp. 5998--6008.

\bibitem{kahneman2011thinking}
Daniel Kahneman,
\newblock {\em Thinking, Fast and Slow},
\newblock Farrar, Straus and Giroux, New York, 2011.

\bibitem{bengio2017consciousness}
Yoshua Bengio,
\newblock ``The consciousness prior,''
\newblock {\em arXiv preprint arXiv:1709.08568}, 2017.

\bibitem{chow1970optimum}
C.~K. Chow,
\newblock ``On optimum recognition error and reject tradeoff,''
\newblock {\em IEEE Trans. Inf. Theory}, vol. 16, no. 1, pp. 41--46, 1970.

\bibitem{geifman2017selective}
Yonatan Geifman and Ran El-Yaniv,
\newblock ``Selective classification for deep neural networks,''
\newblock in {\em Advances in Neural Information Processing Systems (NeurIPS)}, 2017, pp. 4878--4887.

\bibitem{hendrycks2017baseline}
Dan Hendrycks and Kevin Gimpel,
\newblock ``A baseline for detecting misclassified and out-of-distribution examples in neural networks,''
\newblock in {\em Proc. Int. Conf. Learn. Represent. (ICLR)}, 2017.

\bibitem{guo2017calibration}
Chuan Guo, Geoff Pleiss, Yu~Sun, and Kilian~Q. Weinberger,
\newblock ``On calibration of modern neural networks,''
\newblock in {\em Proc. Int. Conf. Mach. Learn. (ICML)}, 2017, pp. 1321--1330.

\bibitem{jaeger2023call}
Paul~F. Jaeger, Carsten~T. L{\"u}th, Lukas Klein, and Till~J. Bungert,
\newblock ``A call to reflect on evaluation practices for failure detection in image classification,''
\newblock in {\em Proc. Int. Conf. Learn. Represent. (ICLR)}, 2023.

\bibitem{mozannar2020consistent}
Hussein Mozannar and David Sontag,
\newblock ``Consistent estimators for learning to defer to an expert,''
\newblock in {\em Proc. Int. Conf. Mach. Learn. (ICML)}, 2020, pp. 7076--7087.

\bibitem{narasimhan2022posthoc}
Harikrishna Narasimhan, Wittawat Jitkrittum, Aditya~Krishna Menon, Ankit~Singh Rawat, and Sanjiv Kumar,
\newblock ``Post-hoc estimators for learning to defer to an expert,''
\newblock in {\em Advances in Neural Information Processing Systems (NeurIPS)}, 2022.

\bibitem{jitkrittum2023confidence}
Wittawat Jitkrittum, Neha Gupta, Aditya~Krishna Menon, Harikrishna Narasimhan, Ankit~Singh Rawat, and Sanjiv Kumar,
\newblock ``When does confidence-based cascade deferral suffice?,''
\newblock in {\em Advances in Neural Information Processing Systems (NeurIPS)}, 2023, pp. 9891--9906.

\bibitem{viola2001rapid}
Paul Viola and Michael Jones,
\newblock ``Rapid object detection using a boosted cascade of simple features,''
\newblock in {\em Proc. IEEE Conf. Comput. Vis. Pattern Recognit. (CVPR)}, 2001, pp. 511--518.

\bibitem{teerapittayanon2016branchynet}
Surat Teerapittayanon, Bradley McDanel, and H.~T. Kung,
\newblock ``{BranchyNet}: Fast inference via early exiting from deep neural networks,''
\newblock in {\em Proc. Int. Conf. Pattern Recognit. (ICPR)}, 2016, pp. 2464--2469.

\bibitem{bolukbasi2017adaptive}
Tolga Bolukbasi, Joseph Wang, Ofer Dekel, and Venkatesh Saligrama,
\newblock ``Adaptive neural networks for efficient inference,''
\newblock in {\em Proc. Int. Conf. Mach. Learn. (ICML)}, 2017, pp. 527--536.

\bibitem{Friston2010}
Karl Friston,
\newblock ``The free-energy principle: a unified brain theory?,''
\newblock {\em Nature Reviews Neuroscience}, pp. 127--138, 2010.

\bibitem{jebraeeli2024koopcon}
Vahid Jebraeeli, Bo~Jiang, Derya Cansever, and Hamid Krim,
\newblock ``Koopcon: A new approach towards smarter and less complex learning,''
\newblock in {\em Proc. IEEE Int. Conf. Image Process. (ICIP)}, 2024.

\bibitem{jebraeeli2025expansive}
Vahid Jebraeeli, Bo~Jiang, Hamid Krim, and Derya Cansever,
\newblock ``Generative expansion of small datasets: An expansive graph approach,''
\newblock in {\em Proc. IEEE Int. Conf. Acoust., Speech, Signal Process. (ICASSP)}, 2025.

\bibitem{khosla2011dogs}
Aditya Khosla, Nityananda Jayadevaprakash, Bangpeng Yao, and Li~Fei-Fei,
\newblock ``Novel dataset for fine-grained image categorization: {Stanford} {Dogs},''
\newblock in {\em Proc. CVPR Workshop on Fine-Grained Visual Categorization (FGVC)}, 2011.

\bibitem{wah2011cub}
Catherine Wah, Steve Branson, Peter Welinder, Pietro Perona, and Serge Belongie,
\newblock ``The {Caltech-UCSD} birds-200-2011 dataset,''
\newblock Tech. {R}ep. CNS-TR-2011-001, California Institute of Technology, 2011.

\bibitem{maji2013aircraft}
Subhransu Maji, Esa Rahtu, Juho Kannala, Matthew Blaschko, and Andrea Vedaldi,
\newblock ``Fine-grained visual classification of aircraft,''
\newblock {\em arXiv preprint arXiv:1306.5151}, 2013.

\bibitem{parkhi2012pets}
Omkar~M. Parkhi, Andrea Vedaldi, Andrew Zisserman, and C.~V. Jawahar,
\newblock ``Cats and dogs,''
\newblock in {\em Proc. IEEE Conf. Comput. Vis. Pattern Recognit. (CVPR)}, 2012, pp. 3498--3505.

\bibitem{deng2009imagenet}
Jia Deng et~al.,
\newblock ``{ImageNet}: A large-scale hierarchical image database,''
\newblock in {\em Proc. IEEE Conf. Comput. Vis. Pattern Recognit. (CVPR)}, 2009, pp. 248--255.

\bibitem{howard2019mobilenetv3}
Andrew Howard et~al.,
\newblock ``Searching for {MobileNetV3},''
\newblock in {\em Proc. IEEE/CVF Int. Conf. Comput. Vis. (ICCV)}, 2019, pp. 1314--1324.

\bibitem{oquab2024dinov2}
Maxime Oquab et~al.,
\newblock ``{DINOv2}: Learning robust visual features without supervision,''
\newblock {\em Trans. Mach. Learn. Res.}, 2024.

\bibitem{hu2022lora}
Edward~J. Hu et~al.,
\newblock ``{LoRA}: Low-rank adaptation of large language models,''
\newblock in {\em Proc. Int. Conf. Learn. Represent. (ICLR)}, 2022.

\bibitem{jeevan2024backbone}
Pranav Jeevan and Amit Sethi,
\newblock ``Which backbone to use: A resource-efficient domain specific comparison for computer vision,''
\newblock {\em arXiv preprint arXiv:2406.05612}, 2024.

\bibitem{kornblith2019better}
Simon Kornblith, Jonathon Shlens, and Quoc~V. Le,
\newblock ``Do better {ImageNet} models transfer better?,''
\newblock in {\em Proc. IEEE/CVF Conf. Comput. Vis. Pattern Recognit. (CVPR)}, 2019, pp. 2661--2671.

\bibitem{hendrycks2019benchmarking}
Dan Hendrycks and Thomas~G. Dietterich,
\newblock ``Benchmarking neural network robustness to common corruptions and perturbations,''
\newblock in {\em Proc. Int. Conf. Learn. Represent. (ICLR)}, 2019.

\bibitem{gulrajani2021search}
Ishaan Gulrajani and David Lopez-Paz,
\newblock ``In search of lost domain generalization,''
\newblock in {\em Proc. Int. Conf. Learn. Represent. (ICLR)}, 2021.

\bibitem{chollet2019measure}
Fran{\c{c}}ois Chollet,
\newblock ``On the measure of intelligence,''
\newblock {\em arXiv preprint arXiv:1911.01547}, 2019.

\bibitem{michaelis2019benchmarking}
Claudio Michaelis et~al.,
\newblock ``Benchmarking robustness in object detection: Autonomous driving when winter is coming,''
\newblock {\em arXiv preprint arXiv:1907.07484}, 2019.

\end{thebibliography}

\end{document}